\documentclass[11pt]{article}

\usepackage[preprint]{acl}

\usepackage{times}
\usepackage{latexsym}

\usepackage[T1]{fontenc}
\usepackage[utf8]{inputenc}
\usepackage{microtype}
\usepackage{inconsolata}
\usepackage{graphicx}

\usepackage{amsmath}
\usepackage{amssymb}
\usepackage{booktabs}
\usepackage{multirow}
\usepackage{xcolor}  
\usepackage{stfloats}
\usepackage{placeins}
\usepackage{listings}  
\title{Evolution or Illusion? Rethinking Evaluation in LLM Evolutionary Search}

\author{
  Tal Oved\thanks{Corresponding author: \texttt{Tal.Oved@ibm.com}} \quad
  Roi Pony \quad
  Oshri Naparstek \quad
  Udi Barzelay \\
  IBM Research
}
\begin{document}
\maketitle

\begin{abstract}
LLM-driven evolutionary search finds programs by launching seeds and iterating each one. Papers report a single budget setting, usually one seed run for a fixed number of iterations, and rank methods from that one point. We show this is not enough. We evaluate three evolutionary search strategies on five optimization tasks, commonly used by papers in the genre to report results. We run the analysis over a full grid of seeds and iterations. Our findings suggest that the best way to split a fixed budget between more seeds (width) and more iterations (depth) changes with the strategy, the task, and the total budget. Furthermore, we observe that the ranking of strategies also changes with the budget. On one task the strategy that looks worst at one seed is best at forty seeds. On another the best number of iterations is well below the value common in practice, so extra depth wastes budget that more seeds would turn into score. We provide a measurement protocol that reports the seeds-by-iterations frontier and practical guidance for using it.
\end{abstract}

\begin{figure}[t]
    \centering
    \includegraphics[width=\columnwidth]{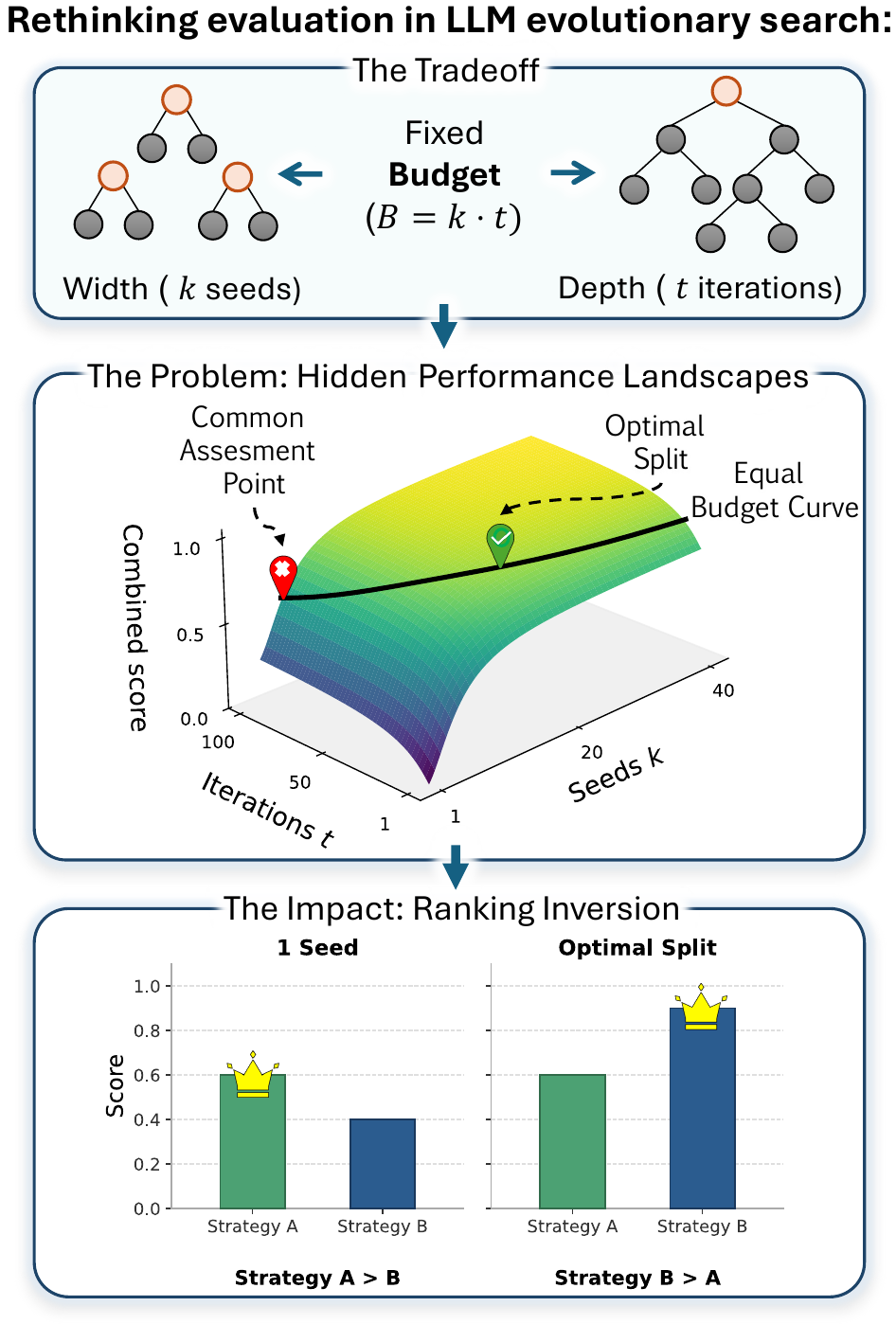}
    \caption{An illustration of the need for rigorous evaluation methodology in LLM-based evolutionary search.}
    \label{fig:my_figure}
\end{figure}

\section{Introduction}

Recent years have seen the emergence of a new paradigm for algorithm discovery, in which large language models generate and iteratively refine candidate algorithms through evolutionary search~\citep{ma2024eureka, fernando2023promptbreeder, guo2024evoprompt, madaan2023self}. FunSearch found new mathematical constructions this way \citep{romera2024funsearch}, AlphaEvolve extended the idea to algorithms and hardware kernels \citep{novikov2025alphaevolve}, and open engines such as OpenEvolve reproduce the loop for public use \citep{openevolve2025}. The execution of these methods is determined by two dimensions, width and depth. Width is the number of independent seeds launched, and the best result over seeds is kept. Depth is the number of iterations each seed runs. Every seed and every iteration spends LLM calls, so a study has a proposal budget $B = k$ seeds $\times\, t$ iterations. A method is usually run for one seed, sometimes up to three, for a fixed number of iterations, and methods are ranked from that single point. The budget is chosen by habit. This paper shows that a single point is not enough to evaluate or compare these methods \citep{henderson2018deep}.

In this work, we address the methodological gap by evaluating three evolutionary search strategies on five optimization tasks over a grid of forty seeds by two hundred iterations. We find that the optimal width-versus-depth split depends on the strategy, task, and budget; that the strategy ranking changes with the budget, so on one task the strategy worst at one seed is best at forty; and that the best number of iterations is often well below common practice, wasting budget more seeds would turn into score.

The message is practical. Report the seeds-by-iterations frontier, not one point. Run several seeds. Tune depth per strategy and task.

\paragraph{Contributions.}
\begin{itemize}
\itemsep0.15em
\item A budget-grid evaluation protocol and practical reporting guidance: replay logged trajectories and compute the expected best score per split $(k,t)$ with exact order statistics (\S\ref{sec:method}, \S\ref{sec:conclusion}).
\item Evidence across three strategies and five tasks that the optimal width-versus-depth split depends on strategy, task, and budget (\S\ref{sec:results}).
\item A ranking inversion: single-seed evaluation reorders the strategies relative to a multi-seed budget, so the common protocol can name the wrong winner (\S\ref{sec:results}).
\end{itemize}

\paragraph{Related work.}
LLM evolutionary search pairs an LLM proposer with an evaluator in an evolutionary loop: FunSearch \citep{romera2024funsearch}, AlphaEvolve \citep{novikov2025alphaevolve}, OpenEvolve \citep{openevolve2025}, GEPA \citep{agrawal2025gepa}, heuristic-design systems \citep{liu2024eoh,ye2024reevo}, sample-efficient variants \citep{lange2025shinka}, and recent work on harness engineering \citep{ishibashi2026harness}. ADRS-Bench collects systems-optimization tasks for this setting \citep{cheng2025barbarians}. Across these methods the budget is fixed in advance and runs are reported singly or in a handful. A parallel line studies how to spend a fixed sampling or search budget \citep{gideoni2026simple,ellis2026don,brown2024monkeys,schaeffer2025powerlaws,kazdan2025passk,snell2024scaling,misaki2025abmcts}, none comparing evolutionary strategies across a seeds-by-iterations budget as we do. Finally, because individual runs are noisy, a best-of-$k$ score inherently scales with $k$ \citep{smith2006optimizer}. We connect these threads: we measure the width-versus-depth split for each strategy and task, and show the reported winner depends on it.

\section{Method}
\label{sec:method}

\paragraph{Setup.}
A strategy runs independent seeds. Seed $i$ produces a trajectory of candidate scores $f(x_{i,1}),\dots,f(x_{i,t})$, where $f$ is the task objective (\texttt{combined\_score}, higher is better). Its value at depth $t$ is the running best
\begin{equation}
M_i(t)=\max_{1\le s\le t} f(x_{i,s}).
\end{equation}
Launching $k$ seeds and keeping the best gives
\begin{equation}
\mathcal{B}_k(t)=\max_{1\le i\le k} M_i(t),
\end{equation}
and the budget is $B=k\,t$.

\begin{figure*}[t]
\centering
\includegraphics[width=\textwidth]{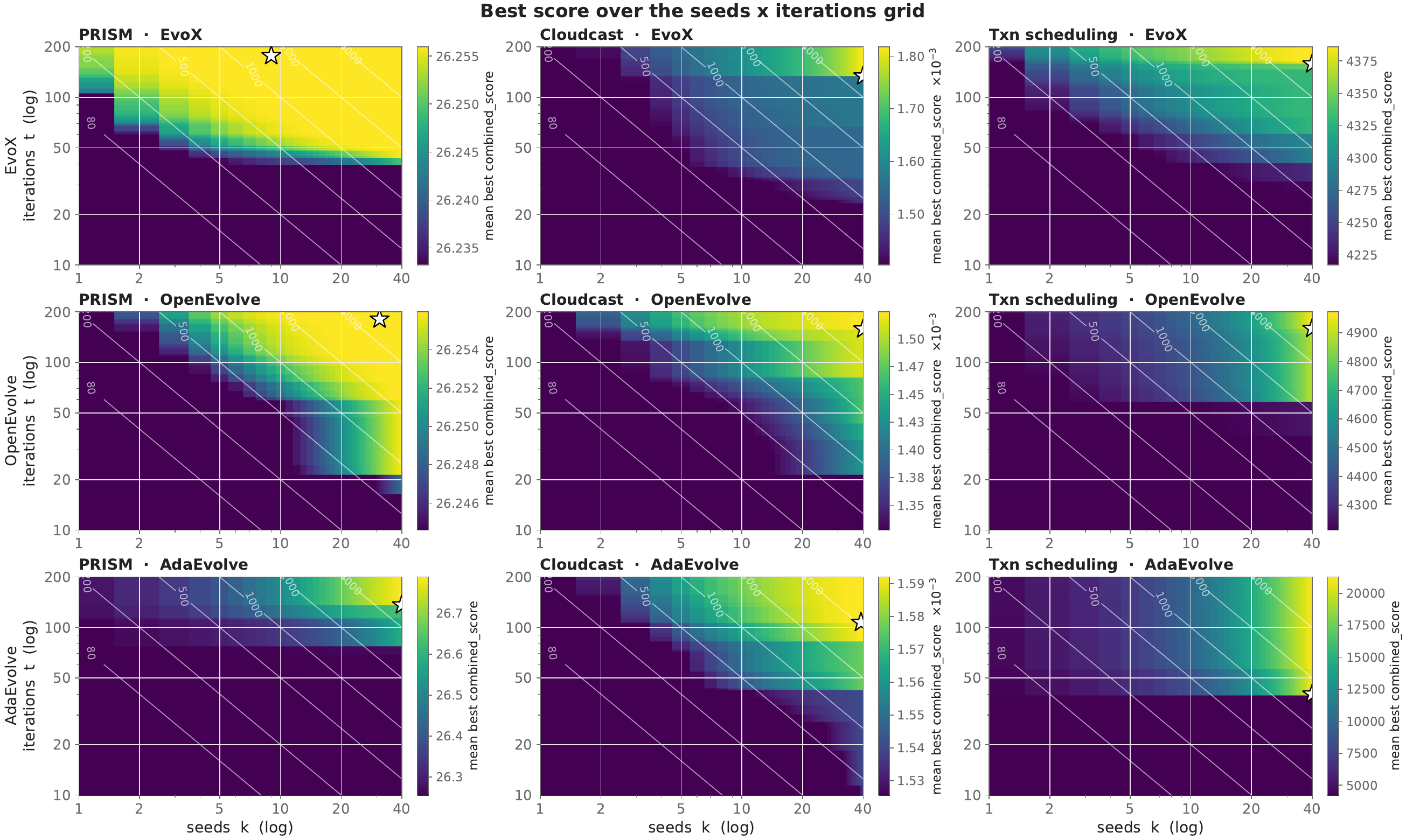}
\caption{Expected best \texttt{combined\_score} over the seeds ($k$) by iterations ($t$) grid (EvoX top, OpenEvolve middle, AdaEvolve bottom). Log-scaled axes make the white iso-budget lines ($k\cdot t$) straight. Stars mark the score-maximizing cells. The star sits in a different place in each panel: the best split depends on the strategy and the task, and the best depth is often well short of the full two hundred iterations.}
\label{fig:heatmap}
\end{figure*}

\paragraph{Budget grid.}
From the logged trajectories of the $n{=}40$ seeds we compute, for every split $(k,t)$, the expected best score $\mathbb{E}\!\left[\mathcal{B}_k(t)\right]$ over random size-$k$ subsets of the $40$ observed seeds (a finite-sample expectation). We use exact order statistics on the per-seed values $\{M_i(t)\}$, so there is no resampling noise. This yields a score surface over seeds and iterations for each strategy and task (Fig.~\ref{fig:heatmap}). We examine three axes:
\begin{itemize}
\itemsep0.1em
\item \textbf{Width marginal}: $\mathbb{E}[\mathcal{B}_k(T)]$ against $k$ at full depth $T{=}200$, the value of more seeds.
\item \textbf{Depth marginal}: $\mathbb{E}[\mathcal{B}_1(t)]$ against $t$ for one seed, the value of more iterations.
\item \textbf{Frontier}: for each total budget $B$, the best expected score over all splits with $k\,t\le B$, and the split $(k,t)$ that reaches it. That split is the recommended allocation at budget $B$.
\end{itemize}
Replaying logged trajectories ensures exact comparisons without requiring redundant runs.

\paragraph{Ranking.}
At a budget point we rank strategies by expected best score. The single-seed protocol common in practice is the point $k{=}1,\,t{=}T$. We compare its ranking to the multi-seed budget $k{=}40,\,t{=}T$.

\paragraph{Ranking reversion.}
We also ask how often a partial budget names the wrong order. We bootstrap the seeds: for each of two thousand resamples we rank the strategies by best-of-seeds score at a budget point and at the full budget, and record whether the orders differ. The reversion probability is the share of resamples that differ, zero at the full budget by construction. We read it over iterations at full width, seeds at full depth, and total budget at its best split.

\section{Experimental Setup}
\label{sec:setup}

\paragraph{Engine and strategies.}
All runs use the ADRS engine \citep{cheng2025barbarians} with \texttt{gpt-5-mini}, keeping the model, objective, and evaluator identical so only the strategy varies. The three strategies are \textbf{OpenEvolve} (islands plus archive, population 40; \citealp{openevolve2025}), \textbf{EvoX} (co-evolving its selection rule, \citealp{liu2026evox}), and \textbf{AdaEvolve} (adaptive variant, \citealp{cemri2026adaevolve}). Following prior work, we budget by proposals rather than raw LLM calls, using an 8{,}000-proposal limit ($40\times200$) to ensure an equal comparison.

\paragraph{Tasks and objective.}
We use three ADRS-Bench system tasks \citep{cheng2025barbarians}: \textbf{PRISM} (LLM-serving scheduler, measured by goodput); \textbf{Cloudcast} (broadcast planner, transfer cost); and \textbf{transaction scheduling} (throughput). We optimize their native \texttt{combined\_score}s (PRISM $\text{reciprocal}(\text{max\_kvpr})+\text{success\_rate}$, transaction scheduling $10^6/(1+\text{makespan})$, Cloudcast cost-based), gating out invalid or reward-hacking candidates via validity checks for comparability. But the transaction-scheduling evaluator is incomplete: it does not verify that a schedule covers the whole workload, so a partial-schedule exploit passes (Appendix~\ref{app:txn-program}); we keep whatever it accepts, as prior work does. PRISM and Cloudcast are deterministic; transaction scheduling draws one random workload per evaluation, so we re-evaluate each seed's best program to remove single-draw noise. We also analyze two math tasks (Appendix~\ref{app:math_tasks}): Circle Packing (Square), and Heilbronn (triangle).

\section{Results}
\label{sec:results}

\paragraph{The optimal split depends on strategy, task, and budget.}
Figure~\ref{fig:heatmap} shows the seeds-by-iterations surface for each strategy and task, with iso-budget lines and the best cell marked. The best cell sits in different places across panels. Table~\ref{tab:split} reports the optimal split at full and 10\% budgets, highlighting two patterns. First, allocations diverge by strategy: on PRISM at 10\% budget, EvoX is best deep and narrow (four seeds, $196$ iterations) while OpenEvolve is best wide and shallow ($36$, $22$). Second, depth saturates early. AdaEvolve on transaction scheduling peaks at $40$ iterations and EvoX on PRISM at $60$; further depth wastes proposals better spent on seeds. Consequently, the optimal allocation axis is highly strategy- and task-specific.

\begin{figure*}[b]
\centering
\includegraphics[width=\textwidth]{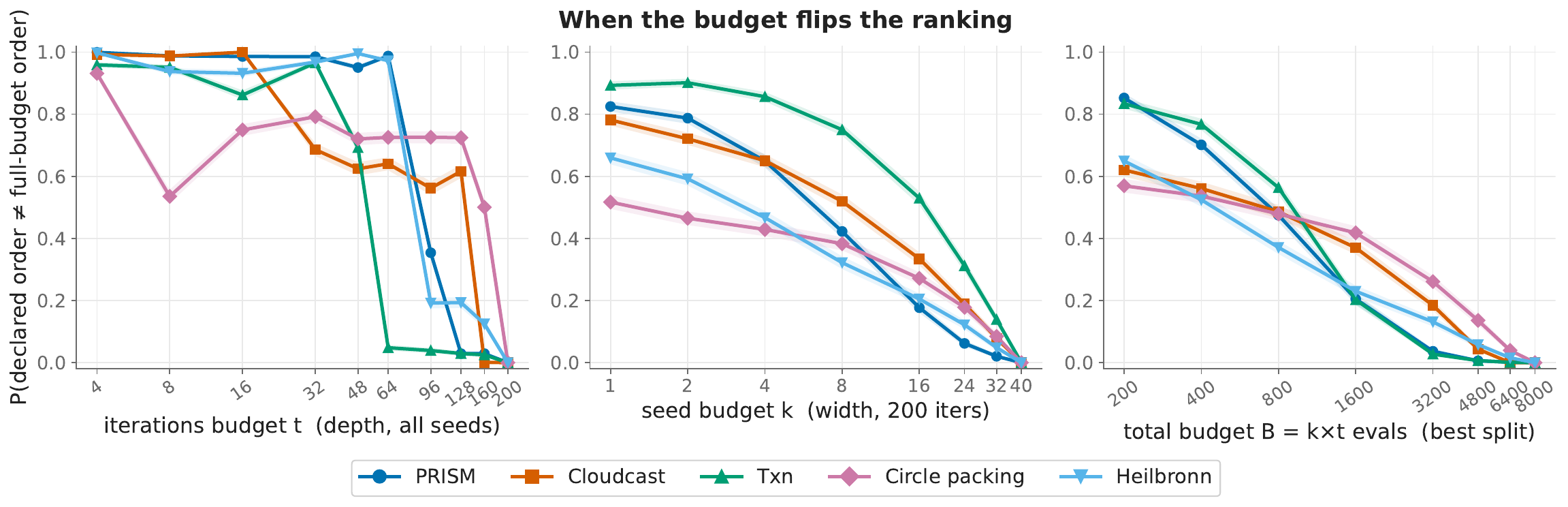}
\caption{Probability that the strategy order declared at a partial budget differs from the full-budget order, one curve per task. Left: iterations at full width. Middle: seeds at full depth. Right: total budget at its best split. Every curve falls to zero as the budget approaches full. The order settles late with depth but steadily with more seeds. Shaded bands are bootstrap $95\%$ confidence intervals.}
\label{fig:reversion}
\end{figure*}

\begin{table}[t]
\centering
\small
\begin{tabular}{llcc}
\toprule
\textbf{Task} & \textbf{Strategy} & \textbf{Full $(k,t)$} & \textbf{10\% $(k,t)$} \\
\midrule
\multirow{3}{*}{PRISM}
 & EvoX       & $(32, 60)$  & $(4, 196)$ \\
 & OpenEvolve & $(32, 171)$ & $(36, 22)$ \\
 & AdaEvolve  & $(40, 136)$ & $(7, 115)$ \\
\midrule
\multirow{3}{*}{Cloudcast}
 & EvoX       & $(40, 134)$ & $(6, 134)$ \\
 & OpenEvolve & $(40, 158)$ & $(5, 162)$ \\
 & AdaEvolve  & $(40, 83)$  & $(7, 114)$ \\
\midrule
\multirow{3}{*}{Txn}
 & EvoX       & $(40, 158)$ & $(4, 198)$ \\
 & OpenEvolve & $(40, 159)$ & $(13, 60)$ \\
 & AdaEvolve  & $(38, 40)$  & $(19, 40)$ \\
\bottomrule
\end{tabular}
\caption{Optimal budget split $(k,t)$  with $kt\le B$ at the full and $\approx$10\% budgets. Allocations vary by strategy and budget. On PRISM at 10\%, prescriptions are opposite (EvoX deep/narrow, OpenEvolve wide/shallow). AdaEvolve on transaction scheduling peaks at $40$ iterations; further depth is wasted.}
\label{tab:split}
\end{table}

\begin{table}[t]
\centering
\small
\begin{tabular}{llcc}
\toprule
\textbf{Task} & \textbf{Strategy} & \textbf{1 seed} & \textbf{40 seeds} \\
\midrule
\multirow{3}{*}{PRISM}
 & EvoX       & $26.254$ (2) & $26.256$ (2) \\
 & OpenEvolve & $26.240$ (3) & $26.256$ (2) \\
 & AdaEvolve  & $\mathbf{26.285}$ (1) & $\mathbf{26.788}$ (1) \\
\midrule
\multirow{3}{*}{\shortstack[l]{Cloudcast\\($\times 10^{-3}$)}}
 & EvoX       & $1.19$ (3) & $\mathbf{1.82}$ (1) \\
 & OpenEvolve & $1.33$ (2) & $1.52$ (3) \\
 & AdaEvolve  & $\mathbf{1.50}$ (1) & $1.59$ (2) \\
\midrule
\multirow{3}{*}{Txn}
 & EvoX       & $4255$ (2) & $4386$ (3) \\
 & OpenEvolve & $4229$ (3) & $4973$ (2) \\
 & AdaEvolve  & $\mathbf{4766}$ (1) & $\mathbf{21277}$ (1) \\
\bottomrule
\end{tabular}
\caption{Expected best \texttt{combined\_score} at one and forty seeds (full depth), with rank in parentheses (higher is better). The order changes with the seed count on every task.}
\label{tab:invert}
\end{table}

\paragraph{The ranking inverts with the budget.}
Table~\ref{tab:invert} gives each strategy at one seed and at forty seeds, both at full depth. On Cloudcast the order reverses: EvoX is last at one seed and first at forty seeds, and OpenEvolve moves from second to last. On transaction scheduling EvoX and OpenEvolve swap the second and third places. On PRISM the single-seed scores are within $0.03$ (a near tie) while at forty seeds AdaEvolve leads by $0.53$. These crossings arise because strategies use budgets differently: added seeds help AdaEvolve far more than EvoX on PRISM, while EvoX gains on both axes on Cloudcast, so single-point measurements observe only one corner of this surface.
Figure~\ref{fig:reversion} turns this into a probability: small budgets yield unreliable rankings; while more depth corrects the order late, more seeds fix it steadily. On every task a large part of the budget is needed before the order is reliable. More seeds also surfaced an AdaEvolve program scoring over $4\times$ any prior result, a validator exploit, not a better scheduler (Appendix~\ref{app:txn-program}); that more seeds expose such flaws is part of our argument.

\section{Conclusion}
\label{sec:conclusion}

Evaluating LLM evolutionary search at one budget point is not enough. The best split between seeds and iterations, and even which method wins, depend on the strategy, the task, and the total budget. We recommend three practices. Report the seeds-by-iterations frontier and the optimal split at each budget, not a single score. Run several seeds, since single-seed rankings are unreliable. Tune depth per strategy and task rather than fixing iterations by habit. All three follow from replaying logged trajectories, so they cost no extra runs.

\newpage
\section*{Limitations}

While this study introduces a rigorous assessment methodology necessary for validating improvements in evolutionary processes, we acknowledge that this protocol imposes a higher computational and financial barrier. Evaluating a full grid of seeds and iterations requires significantly more LLM API calls per assessment than traditional single-point evaluations.  Additionally, our budget-grid protocol is an offline evaluation tool designed to accurately rank methods post-hoc. While it reveals that the optimal width-versus-depth split is task-dependent, a prediction of this optimal split, without first exploring the grid, is still an open question.




\bibliography{reference}

\appendix

\FloatBarrier
\section{Implementation details}
\label{app:impl}

All experiments run on SkyDiscover \citep{liu2026skydiscover}, an open framework for AI-driven algorithmic discovery, using its strategy implementations, task evaluators, and logging. Building on it keeps our pipeline reproducible. 

\section{Budget frontier}
\label{app:frontier}

\begin{figure*}[p]
\centering
\includegraphics[width=\textwidth]{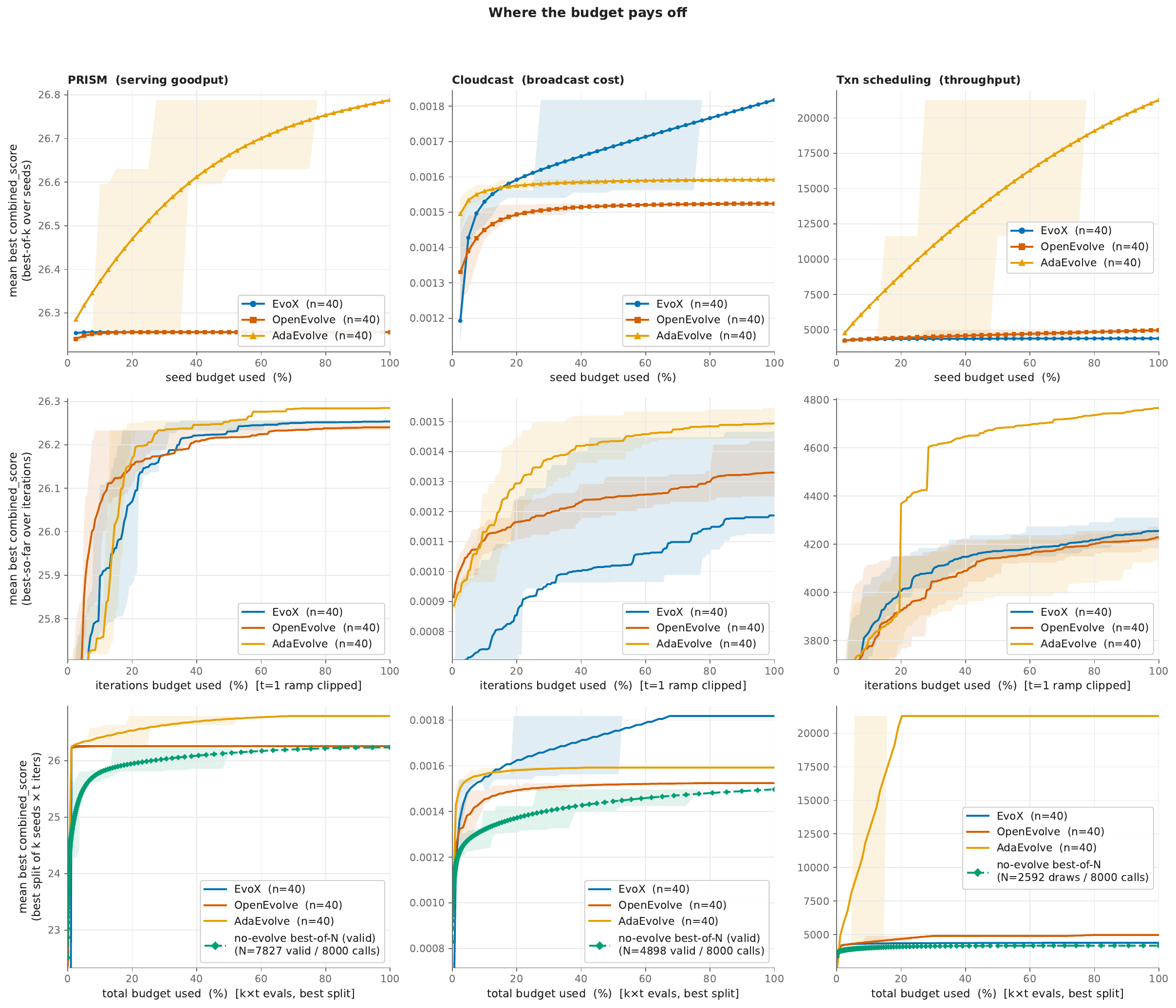}
\caption{Best expected \texttt{combined\_score} against the LLM-call budget, one column per task. Top: value of more seeds at full depth. Middle: value of more iterations for one seed. Bottom: the frontier, the best score at each total budget using the optimal split. The gaps between strategies open and close as the budget grows, so the strategy to prefer depends on how many calls are available. The no-evolve best-of-N line is a reference, not part of the strategy comparison. Shaded bands span the 25th--75th percentile of the best-of-$k$ distribution.}
\label{fig:frontier}
\end{figure*}

Figure~\ref{fig:frontier} plots the best expected score against the total budget, using the optimal
split at each budget. The gaps between strategies open and close as the budget grows, and the leader
changes on some tasks, the effect Figure~\ref{fig:reversion} quantifies.

\begin{figure*}[p]
\centering
\includegraphics[width=\textwidth]{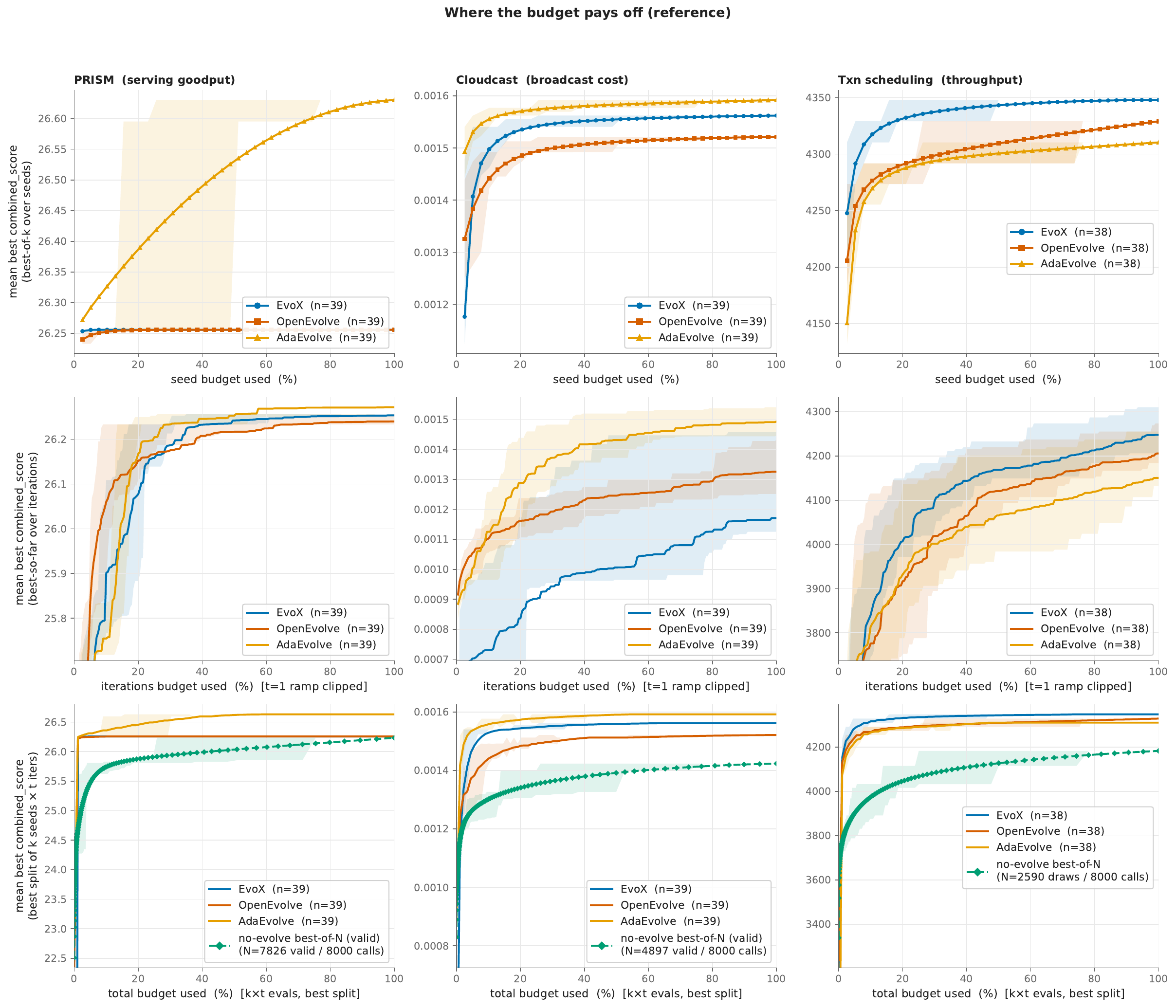}
\caption{Same as Figure~\ref{fig:frontier} with the best seed(s) per technique removed. Trend-reading
only.}
\label{fig:frontier-ref}
\end{figure*}

Figure~\ref{fig:frontier} keeps every seed, so a few lucky seeds set the top of the best-of-k and can
mask the typical trend. Figure~\ref{fig:frontier-ref} repeats it with each technique's best seed
dropped (top two on transaction scheduling, top one on the other tasks, and the best draw of the
no-evolve baseline). Read it for the trend, not for absolute scores.

\newpage
\section{Math tasks}
\label{app:math_tasks}
\begin{figure*}[p]
\centering
\includegraphics[width=\textwidth]{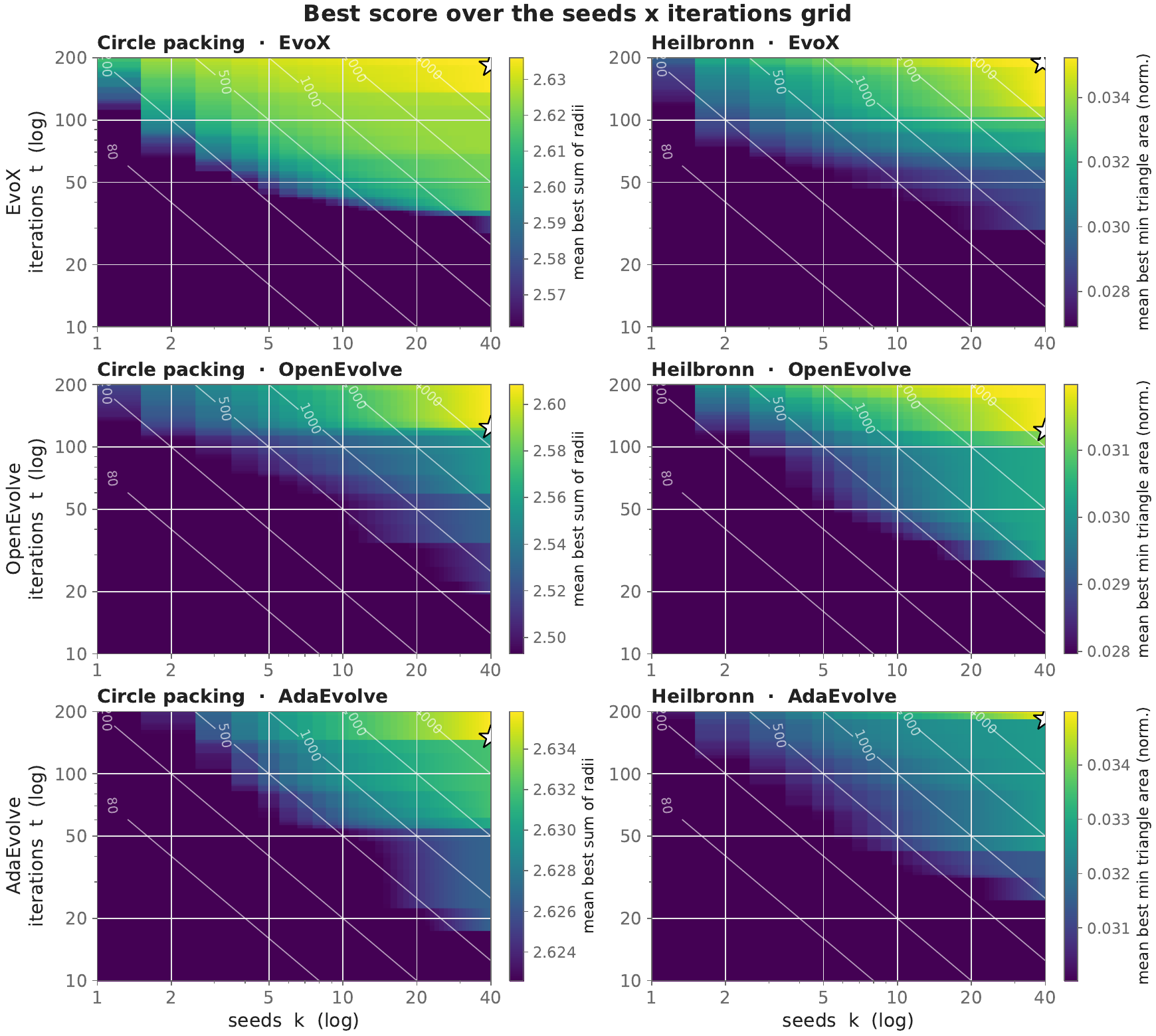}
\caption{Seeds-by-iterations surface for the two math tasks; the best cell is marked. As in the systems tasks, it moves across strategies and depth saturates early.}
\label{fig:heatmap-math}
\end{figure*}

\begin{figure*}[p]
\centering
\includegraphics[width=\textwidth]{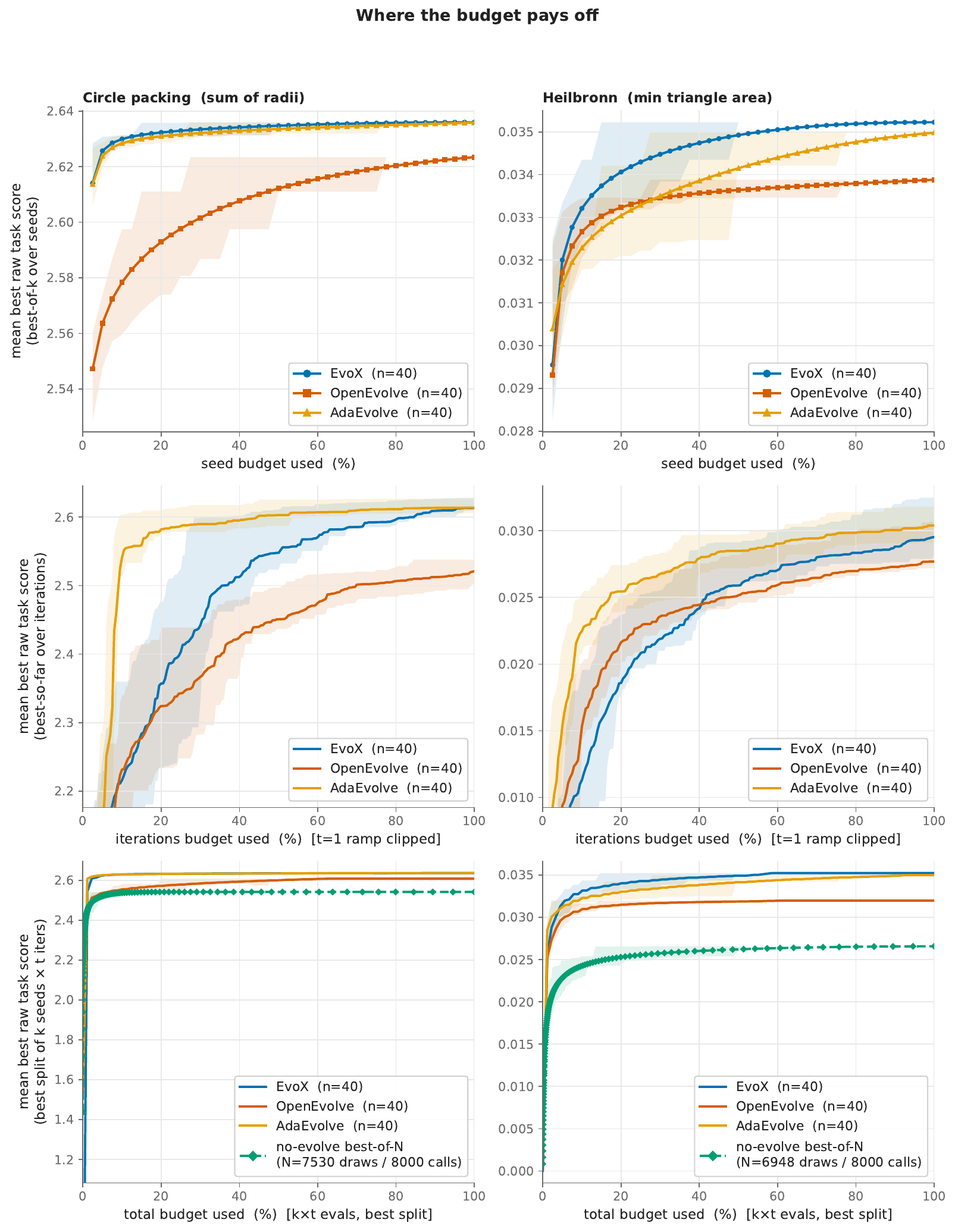}
\caption{Best expected score against the budget for the math tasks. Top: seeds at full depth. Middle: iterations for one seed. Bottom: the frontier at the optimal split.}
\label{fig:frontier-math}
\end{figure*}

The two math tasks show the same budget effects as the systems tasks (Figure~\ref{fig:reversion}). Figure~\ref{fig:heatmap-math} plots the seeds-by-iterations surface: the best cell again lands in different places across strategies, and depth saturates before the full budget. Figure~\ref{fig:frontier-math} shows the frontier rising with the budget, with the gaps between strategies opening and closing as before. Table~\ref{tab:invert-math} confirms the ranking inversion: on Heilbronn, AdaEvolve leads at one seed but EvoX overtakes it at forty, exactly the crossing a single-seed protocol would miss; circle packing demonstrates a near tie between EvoX and AdaEvolve at both budgets. Both tasks support the same conclusion as the systems tasks.

\begin{table}[t]
\centering
\small
\begin{tabular}{llcc}
\toprule
\textbf{Task} & \textbf{Strategy} & \textbf{1 seed} & \textbf{40 seeds} \\
\midrule
\multirow{3}{*}{\shortstack[l]{Circle \\Packing}}
 & EvoX       & $\mathbf{2.6141}$ (1) & $\mathbf{2.6360}$ (1) \\
 & OpenEvolve & $2.5472$ (3) & $2.6234$ (3) \\
 & AdaEvolve  & $2.6137$ (2) & $2.6358$ (2) \\
\midrule
\multirow{3}{*}{\shortstack[l]{Heilbronn\\($\times 10^{-2}$)}}
 & EvoX       & $2.955$ (2) & $\mathbf{3.523}$ (1) \\
 & OpenEvolve & $2.931$ (3) & $3.388$ (3) \\
 & AdaEvolve  & $\mathbf{3.039}$ (1) & $3.498$ (2) \\
\bottomrule
\end{tabular}
\caption{Expected best score at one and forty seeds (full depth), rank in parentheses (higher is better). On Heilbronn, AdaEvolve leads at one seed but EvoX at forty; circle packing is a near tie.}
\label{tab:invert-math}
\end{table}

\newpage
\section{The discovered transaction scheduler}
\label{app:txn-program}

The transaction scheduling column of Table~\ref{tab:invert} looks like our strongest result, but part of it is an artifact of the evaluator. The validity check, \texttt{validate\_schedule}, only tests that the returned sequence is a permutation of the indices it contains. It never checks that every transaction was scheduled. Each workload holds $100$ transactions, so a program can drop some of them, lower its makespan, and still pass with \texttt{validity}$=1.0$.

Running more seeds exposed this gap, and different strategies use it to different degrees (Table~\ref{tab:txncover}). AdaEvolve is the extreme case: its best program scores \texttt{combined\_score} $21277$ (makespan $46$), about $4\times$ any prior result, by scheduling a single transaction per workload and skipping the other $99$. Two of its forty seeds do this. OpenEvolve's best program ($4973$) is subtler: it drops about $45$ of the $100$ transactions in one workload and completes the other two, which is why its makespan ($200$) sits below EvoX's ($227$). Only EvoX schedules every transaction.

We score under the official evaluator to stay comparable with prior work, so we keep every program in the tables and figures. But it means the leaderboard on this task partly reflects how much of the workload a program quietly drops, not how well it schedules. A sound evaluator would need a coverage check, which this benchmark lacks.

We think this makes our point stronger. A one- or three-seed run would rarely reach these programs, while forty seeds surface them, change the ranking, and stress-test the benchmark itself, exposing an evaluator flaw a small budget would miss.

\begin{table*}[t]
\centering
\small
\begin{tabular}{lcccc}
\toprule
\textbf{Strategy} & Best \texttt{combined\_score} & Makespan & Transactions scheduled & Full workload? \\
\midrule
EvoX       & $4386$  & $227$          & $300/300$        & yes \\
OpenEvolve & $4973$  & $\approx\!200$ & $\approx\!255/300$ & no (drops ${\approx}45$) \\
AdaEvolve  & $21277$ & $46$           & $3/300$          & no (drops $297$) \\
\bottomrule
\end{tabular}
\caption{The best transaction-scheduling program found by each strategy at forty seeds, under the official evaluator. Each of the three workloads holds $100$ transactions, for $300$ in total. AdaEvolve and OpenEvolve both pass the validity check while leaving transactions unscheduled, which lowers their makespan and inflates \texttt{combined\_score}; only EvoX schedules the full workload. The evaluator accepts all three because \texttt{validate\_schedule} never checks coverage.}
\label{tab:txncover}
\end{table*}

\end{document}